\documentclass[runningheads]{llncs}
\usepackage{graphicx}
\usepackage{graphicx}
\usepackage{amsfonts,amsmath}
\usepackage{lineno,hyperref}
\usepackage{subfig}
\usepackage{float}
\usepackage{multirow}
\usepackage{bbm}

\newcommand{\tm}{\widetilde{m}}
\newcommand{\one}{\mathbbm{1}}

\begin{document}
\title{Fusion of evidential CNN classifiers for image classification}
\titlerunning{Fusion of evidential CNN classifiers for image classification}
%
\author{Zheng Tong\inst{1}\orcidID{0000-0001-6894-3521} \and
Philippe Xu\inst{1}\orcidID{0000-0001-7397-4808} \and \\
Thierry Den{\oe}ux\inst{1,2}\orcidID{0000-0002-0660-5436}}
\authorrunning{Z. Tong et al.}
%
\institute{Universit\'e de technologie de Compi\`egne, \\
CNRS, UMR 7253 Heudiasyc,  Compi\`egne, France \and
Institut universitaire de France, Paris, France\\
\email{zheng.tong@hds.utc.fr; philippe.xu@hds.utc.fr; thierry.denoeux@utc.fr}}

\maketitle              
\begin{abstract}
We propose an information-fusion approach based on belief functions to combine convolutional neural networks. In this approach, several pre-trained DS-based CNN architectures extract features from input images and convert them into mass functions on different frames of discernment. A  fusion module then aggregates these mass functions using Dempster's rule. An end-to-end learning procedure allows us to fine-tune the overall architecture using a learning set with soft labels, which further improves the classification performance. The effectiveness of this approach is demonstrated experimentally using three benchmark databases.

\end{abstract}
\section{Introduction}
\label{sec:introduction}

Deep learning-based models, especially convolutional neural networks (CNNs) \cite{lecun2015deep} and their variants (see, e.g., \cite{romero2014fitnets}), have been widely used for image classification and have achieved remarkable success. To train such networks, several image data sets are available, with different sets of classes and different granularities. For instance, a dataset may contain images from dogs and cats, while another one may contain images from several species of dogs.  The problem then arises of combining networks trained from such heterogenous datasets. The fusion procedure should be flexible enough to allow the introduction of new datasets with different sets of classes at any stage. 

In this paper, we address this classifier fusion problem in the framework of the Dempster-Shafer (DS) theory of belief functions. The DS theory \cite{shafer1976mathematical}, also known as \emph{evidence theory}, is based on representing independent pieces of evidence by mass functions and combining them using a generic operator called Dempster's rule. DS theory is a well-established formalism for reasoning and making decisions with uncertainty \cite{denoeux20b}. One of its applications is evidential classifier fusion, in which classifier outputs are converted into mass functions and fused by Dempster's rule \cite{quost11,xu2016multimodal}. The information-fusion capacity of DS theory makes it possible to combine deep-learning classifiers.

We present a modular fusion strategy based on DS theory for combining different CNNs. Several pre-trained DS-based CNN architectures, also known as \emph{evidential deep-learning classifiers} in \cite{tong2021evidential}, extract features from input images and convert them to mass functions defined on different frames of discernment. A fusion module then aggregates these mass functions  by Dempster's rule, and the  aggregated mass function is used for classification in a refined frame. An end-to-end learning procedure allows us to fine-tune the overall architecture using a learning set with soft labels, which further improves the classification performance. The effectiveness of the approach is demonstrated experimentally using three benchmark databases: CIFAR-10 \cite{krizhevsky2009learning}, Caltech-UCSD Birds 200 \cite{WelinderEtal2010}, and Oxford-IIIT Pet \cite{parkhi12a}.

The rest of the paper is organized as follows.  DS theory is recalled in Section \ref{sec:background}. The proposed approach is then introduced in Section \ref{sec:framework}, and the numerical experiment is reported in Section \ref{sec:experiment}. Finally, we conclude the paper in Section \ref{sec:conclusion}.

\section{Dempster-Shafer theory}
\label{sec:background}

Let $\Theta=\{\theta_1,\dots,\theta_M\}$ be a set of classes, called the \emph{frame of discernment}. A (normalized) \emph{mass function} on $\Theta$ is a function $m$ from $2^\Theta$ to [0,1] such that $m(\emptyset)=0$ and
$
\sum_{A\subseteq\Theta}m(A)=1.
$
 For any $A\subseteq\Omega$, each mass $m(A)$ is interpreted as a share of a unit mass of belief allocated to the hypothesis that the truth is in $A$, and which cannot be allocated to any strict subset of $A$ based on the available evidence. Set $A$ is called a \emph{focal set} of $m$ if $m(A)>0$. A mass function is \emph{Bayesian} if its focal sets are singletons; it is then equivalent to a probability distribution.

A \emph{refining} from a frame $\Theta$ to a frame $\Omega$, as defined in \cite{shafer1976mathematical}, is a mapping $\rho:  2^\Theta \rightarrow 2^\Omega$ such that the collection of sets $\rho(\{\theta\}) \subset \Omega$ for all $\theta \in \Theta$ form a partition of $\Omega$, and 
 \begin{equation}
 \label{con:refine2}
 \forall A \subseteq \Theta, \quad \rho(A)= \bigcup_{\theta  \in A} \rho(\{\theta\}).
 \end{equation}
The frame $\Omega$ is then called a \emph{refinement} of $\Theta$. Given a mass function $m^\Theta$ on $\Theta$, its vacuous extension $m^{\Theta \uparrow \Omega}$  in $\Omega$ is the mass function defined on  frame $\Omega$ as  
\begin{equation}
\label{con:refine_mass}
m^{\Theta \uparrow \Omega}(B) = \begin{cases}
 m^\Theta(A) \quad & \text{if }   \exists A \subseteq \Theta,\quad B = \rho(A),\\ 
 0 & \rm{otherwise,} 
 \end{cases}
\end{equation}
for all $B \subseteq \Omega$. Two frames of discernment $\Theta_1$ and $\Theta_2$ are said to be \emph{compatible} if they have a common refinement $\Omega$.

Two  mass functions $m_1$ and $m_2$ on the same frame $\Omega$ representing independent items of evidence can be combined conjunctively by \emph{Dempster's rule}  \cite{shafer1976mathematical} defined as follows:
 \begin{equation}
 \label{con:dempster}
 (m_1\oplus m_2)\left(A\right)=\frac{\sum_{B\cap C=A}m_1\left(B\right)m_2\left(C\right)}{\sum_{B\cap C\neq\emptyset}m_1\left(B\right)m_2\left(C\right)}
 \end{equation}
 for all $A\subseteq\Omega$, $A\neq\emptyset$, and $(m_1\oplus m_2)(\emptyset)= 0$. Mass functions $m_1$ and $m_2$ can be combined if and only if the denominator in the right-hand side of \eqref{con:dempster} is positive. The operator $\oplus$ is commutative and associative. The mass function $m_1\oplus m_2$ is called the \emph{orthogonal sum} of $m_1$ and $m_2$. Given two mass functions $m^{\Theta_1}$ and $m^{\Theta_2}$ on compatible frames $\Theta_1$ and $\Theta_2$, their orthogonal sum $m^{\Theta_1}\oplus m^{\Theta_2}$ is defined as the orthogonal sum of their vacuous extensions in their common refinement $\Omega$: $m^{\Theta_1}\oplus m^{\Theta_2}=m^{\Theta_1\uparrow \Omega}\oplus m^{\Theta_2\uparrow \Omega}$.


\section{Proposed approach}
\label{sec:framework}

In this section, we describe the proposed framework for  fusion of evidential deep learning classifiers. The overall architecture is first described in Section \ref{sec:overview}. The end-to-end learning procedure is then introduced in   Section \ref{sec:learning}.

\subsection{Overview}
\label{sec:overview}

\begin{figure}[tb]
 \centering
 \includegraphics[width=\linewidth]{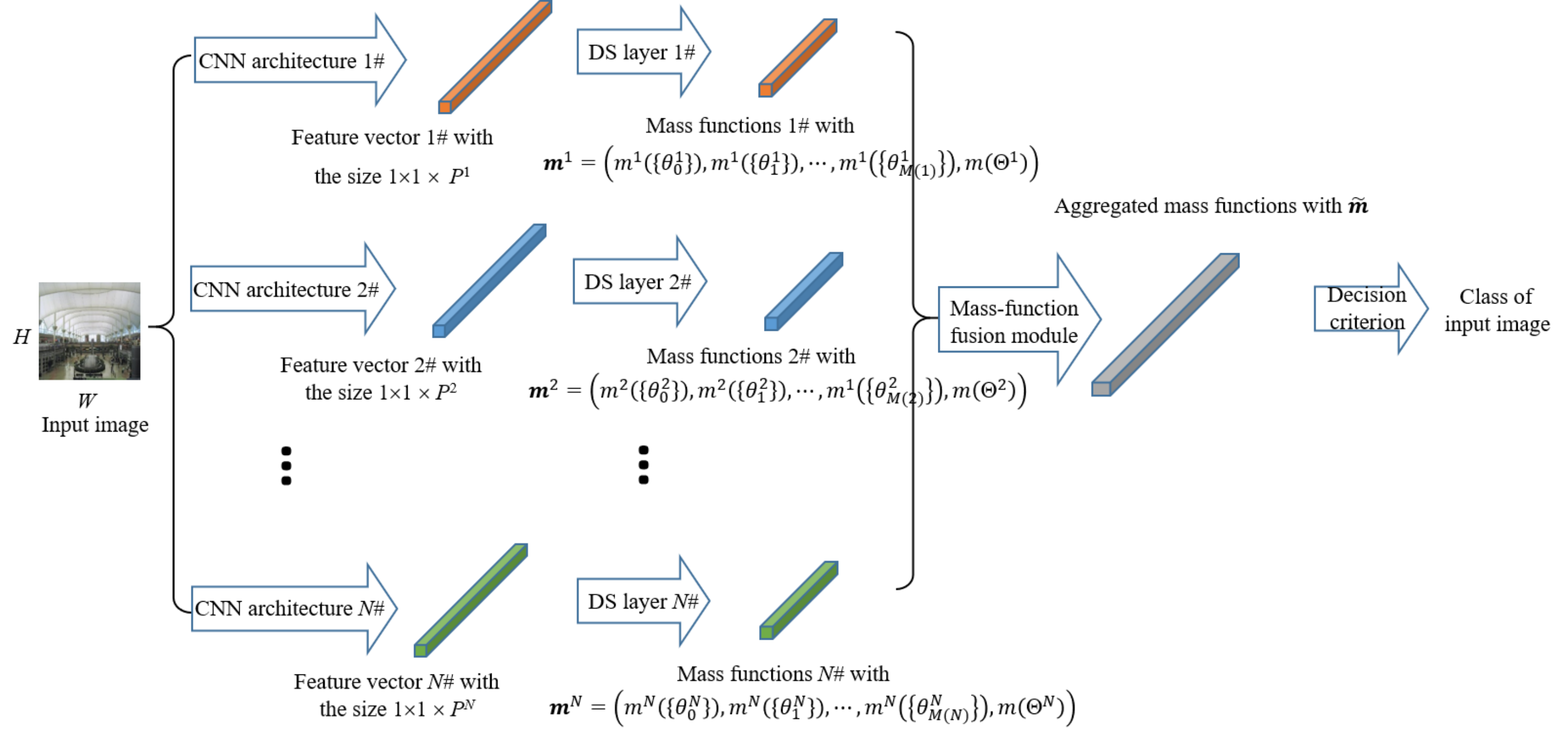}
 \caption{Architecture of a MF-ECNN classifier.}\label{fig:framwork}
\end{figure}

The main idea of the proposed approach is to combine different pre-trained evidential CNN classifiers by plugging a mass-function fusion module at the outputs of these CNN architectures. The architecture of the proposed approach, called \emph{mass-fusion evidential CNN (MFE-CNN) classifier}, is illustrated in Figure \ref{fig:framwork} and can be defined by the following three-step procedure.


\begin{description}
\item{Step 1:} An input image is classified by  $N$ pre-trained DS-based CNN architectures \cite{tong2021evidential}. The $n$-th CNN architecture, $n=1,\dots,N$, extracts a feature vector from the input, as done in a probabilistic CNN  \cite{lecun2015deep}. The vector is then fed into an evidential distance-based neural-network layer for constructing mass functions, called \emph{the DS layer} \cite{denoeux2000neural}. Each unit in this layer computes a mass function on  the frame of discernment $\Theta^n$ composed of $M(n)$ classes $\theta^n_1,\ldots,\theta^n_{M(n)}$ and an ``anything else'' class $\theta^n_0$, based on the distance between the feature vector and a prototype. The mass on $\Theta^n$ is larger when the feature vector is further from the prototype. The mass functions computed by each of the hidden units are then combined by Dempster's rule. Given the design of the DS layer, the focal sets of mass function $m^n$ are the singletons $\{\theta^n_k\}$ for $k=1,\ldots,M(n)$ and $\Theta^n$. 
The outputs after this first step are the $N$ mass functions $m^{1}, \dots,  m^{N}$ defined on $N$ compatible frames $\Theta^1,\dots,\Theta^N$.

\item{Step 2:} A mass-function fusion module aggregates the $N$ mass functions by Dempster's rule. Let $\Omega$ be a common refinement of the $N$ frames $\Theta^1,\dots,\Theta^N$. A combined mass function $\tm$ on $\Omega$ is computed as the orthogonal sum of the $N$ mass functions $\tm=m^1 \oplus \ldots\oplus m^n$. This final output of the mass-function fusion module represents the total evidence about the class of the input image based on the outputs of the $N$ CNN classifiers.

\item{Step 3:} 
The pignistic criterion \cite{smets90e}\cite{denoeux2019decision} is used for decision-making: the mass function $\tm$ is converted into the pignistic probability $BetP_{\tm}$ as
\[
BetP_{\tm}(\{\omega\})=\sum_{\{A \subseteq \Omega : \omega \in A\}} \frac{\tm(A)}{|A|},
\]
for all $\omega\in \Omega$, and the final prediction is $\widehat{\omega}=\arg \max_{\omega \in \Omega} BetP_{\tm}(\omega)$.
\end{description}

\subsection{Learning}
\label{sec:learning}

An end-to-end learning procedure is proposed to fine-tune all the parameters in a MFE-CNN classifier using a learning set with soft labels, in order to improve the classification performance. In the procedure, the learning sets of different pre-trained CNN architectures are merged into a single one. As some labels become imprecise after merging, they are referred to as \emph{soft labels}. For example, the ``bird'' label in the CIFAR-10 \cite{krizhevsky2009learning} database becomes imprecise when the database is merged with the Caltech-UCSD Birds 200 database containing 200 bird species \cite{WelinderEtal2010}. To fine-tune the different classifiers using a learning set with soft labels, we define a label as a non-empty subset  $A \in 2^\Omega \backslash \emptyset$ of classes an image may belong to. Label $A$ indicates that the true class is known to be one element of set $A$, but one cannot determine which one specifically if $|A| > 1$.

In the fine-tuning phase, all parameters in a MFE-CNN classifier are initialized by the parameters in the pre-trained CNNs. Given a learning image with non-empty soft label  $A \subseteq \Omega$ and output pignistic probability $BetP_{\tm}$, we define the loss  as:
\begin{equation}\label{con:loss}
\mathcal{L}(BetP_{\tm}, A)=-\log BetP_{\tm}(A)= -\log \sum_{\omega \in A} BetP_{\tm}(\omega).
\end{equation}
This loss function is minimized when the  pignistic probability of soft label $A$ equals 1. The gradient of this loss w.r.t all network parameters can be back propagated from the output layer to the input layer.

\section{Experiment}
\label{sec:experiment}

In this section, we study the performance of the above fusion method through  a numerical experiment. The databases and metrics are first  introduced in Section \ref{sec:details}. The  results are then discussed in Section \ref{sec:results}.

\subsection{Experimental settings}
\label{sec:details}

Three databases are considered in this experiment: CIFAR-10 \cite{krizhevsky2009learning}, Caltech-UCSD Birds-200-2010 (CUB) \cite{WelinderEtal2010}, and Oxford-IIIT Pet \cite{parkhi12a}. The CIFAR-10 database was pre-split into 50k  training and 10k testing images. For the CUB (6,033 images) and Oxford (7,349 images) databases, we divided each database into training and testing sets with a ratio of about 1:1. The training and testing sets keep the ratio of about 1:1 in each class. In the fine-tuning procedure, the frames of the three databases are refined into a common one, as shown in Table \ref{tab:class_set}. Thanks to the ``anything else'' classes,  the three frames are compatible. After merging the three databases, there are 56,692 training samples and 16,690 testing samples for, respectively,  fine-tuning  and performance evaluation.

\begin{table}[tb]
 \centering
 \caption{Lists of classes in the CIFAR-10, CUB, Oxford databases. The notations $\theta^2_0$ and  $\theta^3_0$ stand for the ``anything else'' class added to the frames of the CUB and Oxford databases.}
 \label{tab:class_set}
 \resizebox{\textwidth}{!}{
  \begin{tabular}{lp{0.85\textwidth}}
   \hline
   Frame        & \multicolumn{1}{c}{Class}                                                                                                                                \\ \hline
   CIFAR-10    & airplane, automobile, bird, cat,  deer, dog, frog, horse, ship, truck.                                                                                 \\ \hline
   CUB        & cardinal, house wren, $\dots$, (200 species of birds), $\theta^2_0$.                                                                                                         \\ \hline
   Oxford     & bengal, boxer, $\dots$,  (37 species of cats and dogs), $\theta^3_0$.                                                                                                        \\ \hline
   Common frame\phantom{a} & airplane, automobile, deer, frog, horse, ship, truck, cardinal, house wren, $\dots$, (200 species of birds), bengal, boxer, $\dots$, (37 species   of cats and dogs). \\ \hline
 \end{tabular}}
\end{table}

For a testing set $T$ with soft labels, the \emph{average error rate}  is defined as
\begin{equation}\label{con:aa}
AE(T)=1-\frac{1}{|T|}\sum_{i \in T} \one_{A(i)}\left(\widehat \omega(i)\right),
\end{equation}
where $A(i)$ is the soft label of sample $i$,  $\widehat \omega(i)$ is the  predicted class, and $\one_{A(i)}$ is the indicator function of set $A(i)$.

The three CNNs used for the three datasets have the same FitNet-4 \cite{romero2014fitnets} architecture with 128 output units. The numbers of prototypes in the DS layers for the CIFAR-10, CUB, and Oxford databases are, respectively, 70, 350, and 80. We compared the proposed classifier to four classifier fusion systems with the same CNN architectures:
\begin{description}
\item[Probability-to-mass fusion (PMF) \cite{xu2016multimodal}:] we feed the feature vector from each CNN $n$ into a softmax layer to generate a  Bayesian mass function on $\Theta^n$.  The  mass functions from the three CNNs are then combined by Dempster's rule. (It should be noted  that  the vacuous extension of each Bayesian mass function in the common refinement $\Omega$ is no longer Bayesian.)
\item[Bayesian-fusion  (BF) \cite{wei2015bayesian}:]  the feature vector from each CNN  is  converted into a Bayesian mass function on the common frame $\Omega$, by equally distributing the mass of a focal set to its elements;
the obtained Bayesian mass functions are combined by Dempster's rule. This procedure is equivalent to  Bayesian fusion. 
\item[Probabilistic feature-combination  (PFC) \cite{nguyen2018deep}:] the three feature vectors are concatenated to form a new vector of length  384, which is fed into a softmax layer to generate the probability distribution on the common frame. 
\item[Evidential feature-combination  (EFC):] feature vectors are concatenated as in the above PFC approach, but the  aggregated vector is fed into a DS layer to generate an output mass function. The dimension of the aggregated vector and the number of prototypes are, respectively, 192 and 400, to obtain  optimal performance of the EFC-CNN classifier.
\end{description}

\subsection{Results}
\label{sec:results}

Table \ref{tab:performance} shows the average test error rates of the evidential and probabilistic classifiers trained from each of the three datasets, as well as the performances of the different fusion strategies (with and without fine tuning) on each individual dataset, and on the union of the three datasets. 

\begin{table}[tb]
\centering
\caption{Average test error rates of different classifiers. ``E2E'' stands for fine-tuned  classifiers. E- and P-FitNit-4 are the evidential and probabilistic CNN classifiers before fusion. The lowest error rates are in bold and second low are underlined.}
\label{tab:performance}	
	\begin{tabular}{clcccc}
		\hline
		& Classifier & CIFAR-10     & CUB          & Oxford       & Overall    \\ \hline
		\multirow{2}{*}{Before fusion} & E-FitNit-4 & 6.4          & 6.6          & 10.2         & -          \\
		& P-FitNit-4 & 6.5          & 7.4          & 10.5         & -          \\ \hline
		\multirow{8}{*}{After fusion}  & MFE        & \underline{5.0}      & 6.6          & 9.9          & \underline{6.4}  \\
		& PMF        & 5.9          & 7.3          & 10.2         & 7.1        \\
		& BF         & 6.2          & 8.9          & 11.1         & 7.7        \\ \cline{2-6} 
		& E2E MFE    & \textbf{4.5} & \underline{6.5}    & \underline{9.8}    & \textbf{6.0} \\
		& E2E PMF    & 5.4          & 7.3          & 10.1         & 6.8        \\
		& E2E BF     & 6.2          & 8.7          & 10.9         & 7.6        \\
		& E2E EFC    & 6.9          & 7.2          & 11.3         & 7.9        \\
		& E2E PFC    & 6.2          & \textbf{6.4} & \textbf{9.7} & 7.0          \\ \hline
	\end{tabular}
\end{table}

Looking at the performance of the MFE strategy, we can see that, after fusion, the error rates on the CIFAR-10 and Oxford databases decrease, but the one on the CUB database does not change. As shown in Table \ref{tab:performance_cifar}, the error rates for the ``cat'', ``dog'', ``bird'' classes on the CIFAR-10 database decrease, but the ones of other classes do not change after fusion. These observations show that the proposed approach makes it possible to combine  CNN classifiers trained from  heterogenous databases to obtain a more general classifier able to recognize classes from any of the databases,  without degrading the performance of the individual classifiers, and sometimes even yielding better results for some classes. 

\begin{table}[tb]
	\centering
	\caption{Test error rates before and after information fusion on CIFAR-10.}
	\label{tab:performance_cifar}
	\begin{tabular}{llcccccccccc}
		\hline
		& Classifier & aero & mobile & bird & cat  & deer & dog  & frog & horse & ship & truck \\ \hline
		\multirow{2}{*}{Before fusion} & E-FitNit-4 & 2.4  & 3.9    & 6.4  & 13.5 & 9.0  & 10.1 & 5.6  & 6.8   & 3.5  & 2.7   \\
		& P-FitNit-4 & 1.6  & 2.6    & 8.7  & 15.7 & 9.6  & 12.5 & 4.2  & 5.3   & 1.9  & 2.6   \\ \hline
		\multirow{3}{*}{After fusion}  & E2E MFE    & 2.2  & 3.9    & 1.9  & 6.3  & 8.5  & 3.9  & 5.5  & 6.5   & 3.5  & 2.7   \\
		& E2E PMF    & 1.6  & 2.5    & 5.0  & 12.8 & 9.0  & 9.2  & 4.2  & 5.3   & 1.8  & 2.6   \\
		& E2E BF     & 1.5  & 2.5    & 8.1  & 14.0  & 9.0  & 11.0 & 4.1  & 5.2   & 1.8  & 2.5   \\ \hline
	\end{tabular}
\end{table}

 Table \ref{tab:examples}, which shows examples of  mass functions computed by the different classifiers, allows us to explain the good performance of the MFE fusion strategy. The first example from the CIFAR-10 database is misclassified using only the mass function from the classifier trained from this dataset, but the decision is corrected after the evidential fusion because the mass function provided by the classifier trained from the CUB data supports some bird species. In contrast, for images misclassified by the CUB mass function, such as the second instance, the other two individual classifiers do not provide any useful information to correct the decisions. Consequently, the classification performance on the CUB database is not improved. Besides, the mass function from the CIFAR-10 classifier sometimes includes useful information for classifying samples into one species of cat or dog, such as the final instance, even though the number of such examples  is small. This phenomenon is responsible for the small change in the performance on the Oxford database.

\begin{table}[tb]
 \caption{Examples of mass functions (MF's) before and after fusion by the MFE strategy. Only some masses before and after fusion are shown for lack of space.}
 \label{tab:examples}
 \resizebox{\textwidth}{!}{
  \begin{tabular}{lllll}
   \hline
   \multirow{2}{*}{Instance/label}              & \multicolumn{3}{c}{Before fusion} & \multicolumn{1}{c}{\multirow{2}{*}{\begin{tabular}[c]{@{}l@{}}MF on $\Omega$\\ after fusion\end{tabular}}}\\ \cline{2-4}
   & MF from CIFAR-10                                 & MF from CUB                              & MF from Oxford                               &                          \\ \hline
   \multirow{4}{*}{\includegraphics[width=0.08\textwidth]{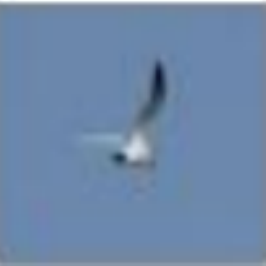}/bird}      & $m(\{\textsf{airplane}\})=0.506$                             & $m(\{\textsf{caspinan}\})=0.698$                    & $m(\{\textsf{samyod}\})=0$                      & $m(\{\textsf{airplane}\})=0.101$                    \\
   & $m(\{\textsf{bird}\})=0.382$                                 & $m(\{\textsf{horned grebe}\})=0.109$                & $m(\{\textsf{pyrenees}\})=0.001$                    & $m(\{\textsf{caspinan}\})=0.672$                    \\
   & $\dots$                                 & $\dots$                & $\dots$                    & $\dots$                   \\
   & $m(\Theta^1)=0.065$ & $m(\theta^2_0)=0.098$                        & $m(\theta^3_0)=0.905$                        & $m(\Omega)=0.007$                          \\ \hline
   \multirow{4}{*}{\includegraphics[width=0.08\textwidth]{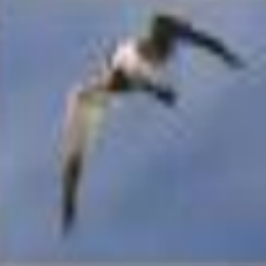}/caspian}  & $m(\{\textsf{airplane}\})=0.009$                             & $m(\{\textsf{caspinan}\})=0.423$                    & $m(\{\textsf{samyod}\})=0$                      & $m(\{\textsf{caspinan}\})=0.415$                    \\
   & $m(\{\textsf{bird}\})=0.823$                                 & $m(\{\textsf{horned grebe}\})=0.452$                & $m(\{\textsf{pyrenees}\})=0.001$                    & $m(\{\textsf{horned grebe}\})=0.450$                \\
   & $\dots$                                 & $\dots$                & $\dots$                    & $\dots$                   \\
   & $m(\Theta^1)=0.092$                                 & $m(\theta^2_0)=0.084$                        & $m(\theta^3_0)=0.951$                        & $m(\Omega)=0.009$                          \\ \hline
   \multirow{4}{*}{\includegraphics[width=0.08\textwidth]{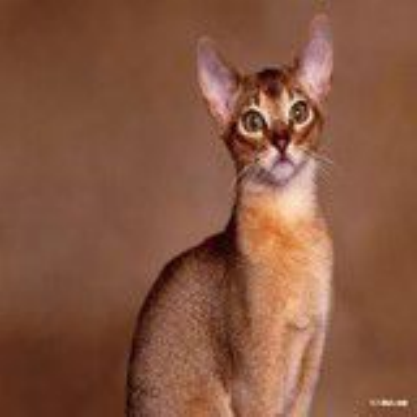}/byssinian} & $m(\{\textsf{cat}\})=0.742$                                  & $m(\{\textsf{caspinan}\})=0.002$                    & $m(\{\textsf{byssinian}\})=0.412$                   & $m(\{\textsf{byssinian}\})=0.414$                   \\
   & $m(\{\textsf{dog}\})=0.131$                                  & $m(\{\textsf{horned grebe}\})=0$                    & $m(\{\textsf{bengal}\})=0.503$                      & $m(\{\textsf{bengal}\})=0.505$                      \\
   & $\dots$                                 & $\dots$                & $\dots$                    & $\dots$                   \\
   & $m(\Theta^1)=0.032$                                 & $m(\theta^2_0)=0.931$                        & $m(\theta^3_0)=0.038$                        & $m(\Omega)=0.005$                          \\ \hline
   \multirow{4}{*}{\includegraphics[width=0.08\textwidth]{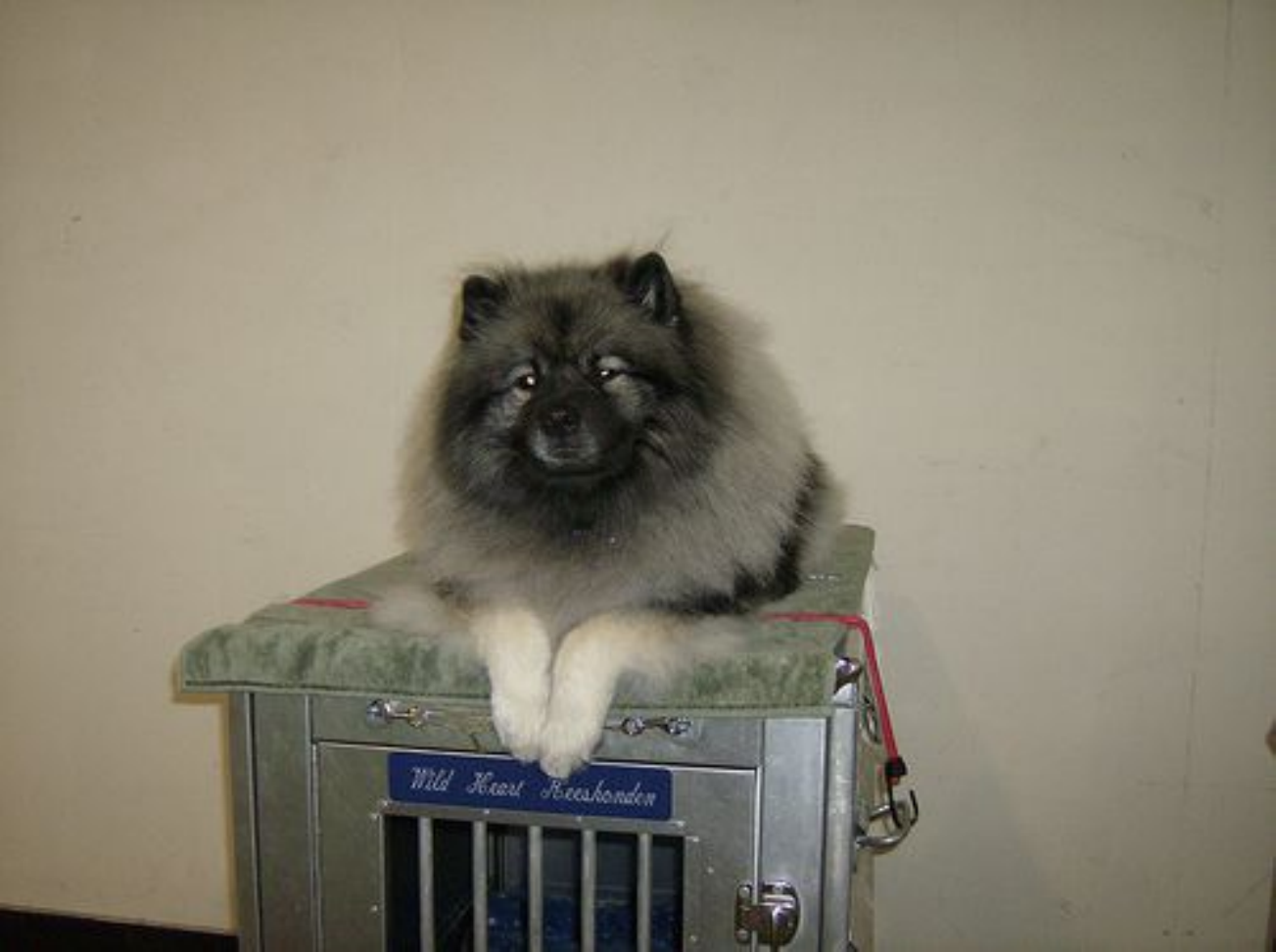}/keeshond} & $m(\{\textsf{cat}\})=0.158$                                  & $m(\{\textsf{albatross}\})=0.001$                    & $m(\{\textsf{rogdoll}\})=0.682$                   & $m(\{\textsf{rogdoll}\})=0.369$                   \\
   & $m(\{\textsf{dog}\})=0.705$                                  & $m(\{\textsf{horned grebe}\})=0$                    & $m(\{\textsf{keeshond}\})=0.254$                      & $m(\{\textsf{keeshold}\})=0.485$                      \\
   & $\dots$                                 & $\dots$                & $\dots$                    & $\dots$                   \\
   & $m(\Theta^1)=0.058$                                 & $m(\theta^2_0)=0.975$                        & $m(\theta^3_0)=0.001$                        & $m(\{\textsf{cat}\})=0.021$                          \\ \hline
  \end{tabular}
 }
\end{table}

Comparing the test error rates of the MFE classifiers with and without the end-to-end learning procedure as shown in Table \ref{tab:performance} , we can see that fine tuning further boosts the overall performance, as well as the performance on the CIFAR-10 and Oxford databases. Thus, the fine-tuning procedure decreases the classification error rate of the proposed architecture, and can be seen as a way to improve the performance of CNN classifiers. This is because the end-to-end learning procedure adapts  the individual classifiers to the new classification problem. More specifically, before fusion, the  CNN classifiers are pre-trained for the classification tasks with the frames of discernment before refinement. 
The proposed end-to-end learning procedure fine-tunes the parameters in the CNN  and DS layers to make them more suitable to the classification task in the refined frame.

Finally, Table \ref{tab:performance}   sheds some light on the relative performance of different classifier fusion strategies.  The PMF fusion strategy also improves the performance of the probabilistic CNNs trained on each of the three databases, but it is not as good as the proposed method. 
In contrast, the Bayesian fusion strategy (BF)  degrades the performance of the individual classifiers, which shows that the method is not effective when the numbers of classes in the different frames are very unbalanced. The relatively high error rates obtained of the two feature fusion strategies (E2E EFC and E2E PFC) show that this simple fusion method is less effective than the other ones. All in all, the proposed evidential fusion strategy outperforms the other tested methods on the datasets considered in this experiment.

\section{Conclusion}
\label{sec:conclusion}

In the study, we have proposed a fusion approach based on belief functions to combine different CNNs for image classification. The proposed approach makes it possible to combine CNN classifiers trained from  heterogenous databases with different sets of classes. The combined classifier is able to classify images from any of these databases while having  at least as good performance as those of the individual classifiers on their respective databases. Besides, the proposed approach makes it possible to combine additional classifiers trained from new datasets with different sets of classes at any stage. An end-to-end learning procedure further improves the performance of the proposed architecture. This approach was shown to outperform  other decision-level or feature-level  fusion strategies for combining CNN classifiers. Future work will consider  combining evidential fully convolutional networks for pixel-wise semantic segmentation \cite{tong2021evidentialfcn}.

%
%
%
\bibliographystyle{splncs04}
\bibliography{mybibliography}

\begin{thebibliography}{10}
\providecommand{\url}[1]{\texttt{#1}}
\providecommand{\urlprefix}{URL }
\providecommand{\doi}[1]{https://doi.org/#1}

\bibitem{denoeux2000neural}
Denoeux, T.: A neural network classifier based on {Dempster-Shafer} theory.
  IEEE Transactions on Systems, Man, and Cybernetics-Part A: Systems and Humans
   \textbf{30}(2),  131--150 (2000)

\bibitem{denoeux2019decision}
Den{\oe}ux, T.: Decision-making with belief functions: a review. International
  Journal of Approximate Reasoning  \textbf{109},  87--110 (2019)

\bibitem{denoeux20b}
Den{\oe}ux, T., Dubois, D., Prade, H.: Representations of uncertainty in
  artificial intelligence: Beyond probability and possibility. In: A Guided
  Tour of Artificial Intelligence Research, vol.~1, chap.~4, pp. 119--150.
  Springer Verlag (2020)

\bibitem{krizhevsky2009learning}
Krizhevsky, A., Hinton, G.: Learning multiple layers of features from tiny
  images. Tech. rep., University of Toronto (2009)

\bibitem{lecun2015deep}
LeCun, Y., Bengio, Y., Hinton, G.: Deep learning. nature  \textbf{521}(7553),
  436--444 (2015)

\bibitem{nguyen2018deep}
Nguyen, L.D., Lin, D., Lin, Z., Cao, J.: Deep {CNNs} for microscopic image
  classification by exploiting transfer learning and feature concatenation. In:
  2018 IEEE International Symposium on Circuits and Systems (ISCAS). pp.~1--5.
  IEEE (2018)

\bibitem{parkhi12a}
Parkhi, O.M., Vedaldi, A., Zisserman, A., Jawahar, C.V.: Cats and dogs. In:
  IEEE Conference on Computer Vision and Pattern Recognition. Providence, Rhode
  Island (2012)

\bibitem{quost11}
Quost, B., Masson, M.H., Den{\oe}ux, T.: Classifier fusion in the
  {Dempster-Shafer} framework using optimized t-norm based combination rules.
  International Journal of Approximate Reasoning  \textbf{52}(3),  353--374
  (2011)

\bibitem{romero2014fitnets}
Romero, A., Ballas, N., Kahou, S.E., Chassang, A., Gatta, C., Bengio, Y.:
  Fitnets: Hints for thin deep nets. In: 3rd International Conference on
  Learning Representations. San Diego, USA (2015)

\bibitem{shafer1976mathematical}
Shafer, G.: A mathematical theory of evidence. Princeton University Press,
  Princeton (1976)

\bibitem{smets90e}
Smets, P.: Constructing the pignistic probability function in a context of
  uncertainty. In: Henrion, M., Schachter, R.D., Kanal, L.N., Lemmer, J.F.
  (eds.) Uncertainty in Artificial Intelligence 5, pp. 29--40. North-Holland,
  Amsterdam (1990)

\bibitem{tong2021evidential}
Tong, Z., Xu, P., Denoeux, T.: An evidential classifier based on
  {Dempster-Shafer} theory and deep learning. Neurocomputing  \textbf{450},
  275--293 (2021)

\bibitem{tong2021evidentialfcn}
Tong, Z., Xu, P., Den{\oe}ux, T.: Evidential fully convolutional network for
  semantic segmentation. Applied Intelligence  (2021),
  \url{https://doi.org/10.1007/s10489-021-02327-0}

\bibitem{wei2015bayesian}
Wei, Q., Dobigeon, N., Tourneret, J.Y.: Bayesian fusion of multi-band images.
  IEEE Journal of Selected Topics in Signal Processing  \textbf{9}(6),
  1117--1127 (2015)

\bibitem{WelinderEtal2010}
Welinder, P., Branson, S., Mita, T., Wah, C., Schroff, F., Belongie, S.,
  Perona, P.: {Caltech-UCSD Birds 200}. Tech. Rep. CNS-TR-2010-001, California
  Institute of Technology (2010)

\bibitem{xu2016multimodal}
Xu, P., Davoine, F., Bordes, J.B., Zhao, H., Den{\oe}ux, T.: Multimodal
  information fusion for urban scene understanding. Machine Vision and
  Applications  \textbf{27}(3),  331--349 (2016)

\end{thebibliography}

\end{document}